\documentstyle[11pt,times,latex-acl,newapa,ipamacs,epic,eepic]{article}

\title{A LEXICAL DATABASE TOOL FOR\\[\smallskipamount]%
  QUANTITATIVE PHONOLOGICAL RESEARCH}
\author{%
  \begin{tabular}{@{\extracolsep{5ex}}ll}
    \multicolumn{2}{c}{Steven Bird} \\[\smallskipamount]
    The University of Edinburgh   & SIL Cameroon\\
    Centre for Cognitive Science  & B.P. 1299\\
    Edinburgh EH8 9LW, UK         & Yaound\'e, Cameroon \\[\smallskipamount]
    \multicolumn{2}{c}{\texttt{Steven.Bird@ed.ac.uk}}
  \end{tabular}
}

\newcommand{\mbaru}{\diatop[\=|\baru]}

\font\ipatwelverm=wsuipa12

\font\ipanine=wsuipa9
\def\normalipa{\def\ipa{\ipatwelverm}}
\def\smallipa{\def\ipa{\ipanine}}
\normalipa

\def\HTML{{\sc html}}

\def\spc{\vspace{1ex}}
\newenvironment{dpy}{\setlength{\parindent}{0pt}%
  \begin{figure*}\begin{minipage}{\textwidth}}{\spc\hrule\end{minipage}\end{figure*}}

\newenvironment{dpycol}{\setlength{\parindent}{0pt}%
  \begin{figure}\begin{minipage}{\linewidth}}{\spc\hrule\end{minipage}\end{figure}}

\newenvironment{query}{\begin{minipage}[t]{\linewidth}%
  {\bf Search Attributes:}\small\tt \begin{center}\begin{tabular}{rl}}%
  {\end{tabular}\end{center}\end{minipage}\vspace{1em}\normalfont\normalsize}

\newenvironment{results}[1]{\begin{minipage}[t]{\linewidth}%
  {\bf Search Results:}\noindent\begin{center}\begin{tabular}{#1}}%
  {\end{tabular}\end{center}\end{minipage}\vspace{1em}}

\newbox{\speechgif}
\savebox{\speechgif}{
\setlength{\unitlength}{0.0001in}
{
\begin{picture}(1224,1239)(0,-10)
\texture{0 115111 51000000 444444 44000000 151515 15000000 444444 
    44000000 511151 11000000 444444 44000000 151515 15000000 444444 
    44000000 115111 51000000 444444 44000000 151515 15000000 444444 
    44000000 511151 11000000 444444 44000000 151515 15000000 444444 }%
\shade\path(12,1212)(1212,1212)(1212,12)(12,12)(12,1212)
\path(12,1212)(1212,1212)(1212,12)(12,12)(12,1212)
\texture{44555555 55aaaaaa aa555555 55aaaaaa aa555555 55aaaaaa aa555555 55aaaaaa 
    aa555555 55aaaaaa aa555555 55aaaaaa aa555555 55aaaaaa aa555555 55aaaaaa 
    aa555555 55aaaaaa aa555555 55aaaaaa aa555555 55aaaaaa aa555555 55aaaaaa 
    aa555555 55aaaaaa aa555555 55aaaaaa aa555555 55aaaaaa aa555555 55aaaaaa }%
\shade\path(162,762)(162,462)(387,462)
    (687,162)(687,1062)(387,762)(162,762)(162,762)
\path(162,762)(162,462)(387,462)(687,162)(687,1062)(387,762)(162,762)(162,762)
\path(762,912)(781.373,853.095)(797.449,802.063)(810.450,758.026)
    (820.594,720.105)(833.191,659.094)(837.000,612.000)
\path(837,612)(833.191,564.906)(820.594,503.895)(810.450,465.974)
    (797.449,421.937)(781.373,370.905)(762.000,312.000)
\path(912,1062)(942.336,1020.290)(970.118,981.159)(1018.348,909.982)
    (1057.349,847.148)(1087.781,791.340)(1110.303,741.240)
    (1125.574,695.528)(1134.253,652.888)(1137.000,612.000)
\path(1137,612)(1134.253,571.112)(1125.574,528.472)(1110.303,482.760)
    (1087.781,432.660)(1057.349,376.852)(1018.348,314.018)
    (970.118,242.841)(942.336,203.710)(912.000,162.000)
\path(837,987)(875.745,916.689)(907.898,855.271)(933.899,801.648)
    (954.188,754.721)(969.202,713.392)(979.383,676.560)(987.000,612.000)
\path(987,612)(979.383,547.440)(969.202,510.608)(954.188,469.279)
    (933.899,422.352)(907.898,368.729)(875.745,307.311)(837.000,237.000)
\end{picture}
}}
\def\speech{\usebox{\speechgif}}

\DeclareFixedFont{\IPAfont}{OT1}{cmr}{m}{n}{12pt}
\def\data#1{{\IPAfont #1}}
\DeclareFixedFont{\smallIPAfont}{OT1}{cmr}{m}{n}{9pt}
\def\smalldata#1{{\smallIPAfont\smallipa #1}}

\begin{document}
\maketitle

\begin{abstract}
  A lexical database tool tailored for phonological res\-earch is
  described.  Database fields include transcriptions, glosses and
  hyperlinks to speech files.  Database queries are expressed using
  \HTML\ forms, and these permit regular expression search on any
  combination of fields.  Regular expressions are passed directly to
  a Perl CGI program, enabling the full flexibility of Perl extended
  regular expressions.  The regular expression notation is extended to
  better support phonological searches, such as search for minimal pairs.
  Search results are presented in the form of
  \HTML\ or \LaTeX\ tables, where each cell is either a number (representing
  frequency) or a designated subset of the fields.  Tables have up
  to four dimensions, with an elegant system for specifying which
  fragments of which fields should be used for the row/column labels.
  The tool offers several advantages over traditional methods
  of analysis: (i) it supports a quantitative method of doing
  phonological research; (ii) it gives universal access to the same
  set of informants; (iii) it enables other researchers to hear the
  original speech data without having to rely on published
  transcriptions; (iv) it makes the full power of regular expression search
  available, and search results are full multimedia documents; and
  (v) it enables the early refutation of false hypotheses,
  shortening the analysis-hypothesis-test loop.
  A life-size application to an African tone language (Dschang)
  is used for exemplification throughout the paper.  The database contains
  2200 records, each with approximately 15 fields.  Running on
  a PC laptop with a stand-alone web server, the
  `Dschang HyperLexicon' has already been used extensively in
  phonological fieldwork and analysis in Cameroon.
\end{abstract}

\bibliographystyle{newapa}

\section{INTRODUCTION}

Initial stages of phonological analysis typically focus on words in
isolation, as the phonemic inventory and syllable canon are
established.
Data is stored as a lexicon, where each word is entered
as a transcription accompanied
by at least a gloss (so the word can be elicited again) and the
major syntactic category.
In managing a lexicon, the working phonologist has a variety
of computational needs:
storage and retrieval;
searching and sorting;
tabular reports on distributions and contrasts;
updates to database and to reports
as distinctions are discovered or discarded.
In the past the analyst had to do all this computation
by hand using index cards kept in shoeboxes.  But now
many of these particular tasks are automated by software such as
the SIL programs Shoebox \shortcite{Shoebox} and Findphone
\cite{Findphone},\footnote{Unlike regular database management systems,
  these include
  international and phonetic character sets and user-defined keystrokes
  for entering them, and a utility to dump a database into an RTF file
  in a user-defined lexicon format for use in desktop publishing.
}
or using commercial database packages.

Of course, many tasks other than those listed above have already
benefitted from (partial) automation.\footnote{For example,
see \cite{Ellison92b,LoweMazaudon94,ColemanDirksen96}.}
Additionally,
it has been shown how a computational inheritance model
can be used for structuring
lexical information relevant for phonology \cite{Reinhard91}.
And there is a body of work on the use of finite
state devices -- closely related to regular expressions --
for modelling phonological phenomena \cite{KaplanKay94} and
for speech processing (cf.\ Kornai's work with HMMs \cite{Kornai95}).
However, computational phonology is yet to provide tools
for manipulating lexical and speech data using the full
expressive power of the regular expression notation in a way that
supports pure phonological research.

This paper describes a lexical database system tailored to
the needs of phonological research and exemplified for Dschang, a
language of Cameroon.  An online lexicon (originally published as
Bird \& Tadadjeu, 1997\nocite{BirdTadadjeu97}), contains records with the format
in Figure~\ref{fig:format}.  Only the most important fields are shown.

\newlength{\colwidth}
\setlength{\colwidth}{0.6\linewidth}
\begin{dpycol}\begin{center}\begin{tabular}{llp{\colwidth}}
\verb|\id| & 1612           & \it identifier (used for hyperlinks)\\
\verb|\v|  &                & \it validation status\\
\verb|\w|  & \data{mbh\mbaru}      & \it orthographic form\\
\verb|\as| & \#m.bhU\#      & \it ascii transcription\\
\verb|\rt| & \#bhU\#        & \it transcription of word root\\
\verb|\t|  & LDH            & \it tone transcription\\
\verb|\sd| & \data{mbh\mbaru}      & \it southern dialect form\\
\verb|\pg| & \data{*b\'u$^+$\`a}   & \it Proto-Grassfields form\\
\verb|\p|  & n              & \it part of speech\\
\verb|\pl| & \data{me-}     & \it plural prefix\\
\verb|\cl| & 9/6            & \it noun class (singular/plural)\\
\verb|\en| & dog            & \it english gloss\\
\verb|\fr| & chien          & \it\raggedright french gloss (used with informants)
\end{tabular}\end{center}
\caption{Format of Database Records}\label{fig:format}
\end{dpycol}

The user interface is provided by a Web browser.  A suite of Perl
programs \cite{Wall91} generates the search form in
\HTML\ and processes the query.  Regular
expressions in the query are passed directly to Perl,
enabling the full flexibility of Perl extended regular expressions.
A further extension to the notation
allows searches for minimal sets, groups of words which are minimally
different according to some criterion.
Hits are structured into a tabular display and returned as an
\HTML\ or \LaTeX\ document.

In the next section, a sequence of example queries is given
to illustrate the format of queries and results, and to demonstrate
how a user might interact with the system. 
A range of more powerful queries are then demonstrated,
along with an explanation of the notations for minimal pairs and projections.
Next, some implementation details are given, and the
component modules are described in detail.
The last two sections describe
planned future work and present the conclusions.

\section{EXAMPLE}
\label{sec:example}

This section shows how the system can be used to support phonological
analysis.  The language data comes from Dschang, a Grassfields Bantu
language of Cameroon, and is structured into a lexicon consisting
of 2200 records.
Suppose we wished to learn about phonotactic constraints
in the syllable rhyme.  The following sequence of queries were not
artificially constructed, but were issued in an actual session with
the system in the field, running the Web server in a stand-alone mode.
The first query is displayed below.\footnote{
  The display is only a crude approximation to the \HTML\ form.
  Note that the query form comes with the variables already filled in
  so that it is not necessary for the user to supply them, although
  they can be edited.
  The transcription symbols used in the system have the following
  interpretation:
  U=\smalldata{\baru}, @=\smalldata{\schwa},
  E=\smalldata{\niepsilon}, O=\smalldata{\openo}, N=\smalldata{\eng},
  '=\smalldata{\glotstop}.
}\spc

\begin{query}
  display:   & count \\
  root:      & .*([\$V])([\$C])\# \\
  loanwords: & exclude \\
  suffixed:  & include \\
  phrases:   & exclude \\
  time-limit:& 2 minutes\\
       vars: & \$B = "$\backslash$.\#-"; \# boundaries\\
             & \$S = "pbtdkgcj'";        \# stops\\
             & \$F = "zsvfZS";           \# fricatives\\
             & \$O = \$S.\$F;            \# obstruents\\
             & \$N = "mnN";              \# nasals\\
             & \$G = "wy";               \# glides\\
             & \$C = \$O.\$N.\$G."hl";   \# cons\\
             & \$V = "ieaouEOU@";        \# vowels
\end{query}\spc

\noindent
The main attribute of interest is the \verb|root|
attribute.\footnote{In the following discussion, `attribute' refers to
  a line in the query form while `field' refers to part of a database
  record.}
The \verb|.*| expression stands for a sequence of zero or more
segments.  The expressions \verb|$V| and \verb|$C| are variables
defined in the \verb|vars| section of the query form.
These are strings, but when surrounded with brackets, as in
\verb|[$V]| and \verb|[$C]|, they function as wild cards which match
a single element from the string.  The \verb|#| character is a
boundary symbol marking the end of the root.
Observe that the \verb|root|
attribute contains two parenthesised subexpressions.  These
will be called {\em parameters} and have a special role in
structuring the search output.  This is best demonstrated by
way of an example.  Consider the table below, which is the
result of the above query.  In this table, the row labels are all the
segments which matched the variable \verb|$V|, while the column labels
are just the segments that matched \verb|$C|.\spc

\begin{results}{c|cccccc}
  & p  & t  & k  & '   & m  & N \\ \hline
i & 5  &    & 10 & 24  & 9  & 32 \\
U & 9  &    &    & 38  & 1  & 9 \\
u & 14 &    &    & 60  & 10 & 39 \\
@ &    &    & 15 &     & 41 & 75 \\
o &    &    & 31 & 12  &    & \\
E & 51 & 14 &    &     &    & \\
a & 30 & 1  & 46 & 61  &    & 76 \\
O & 15 & 1  & 12 & 36  &    & 49
\end{results}\spc

\noindent
There are sufficient gaps
in the table to make us wonder if all the segments are actually
phonemes.  For example, consider \data{o} and \data{u}, given that they are
phonetically very similar
([\data{\closedniomega}] and [\data{u}] respectively).
We can easily set up \data{o}
as an allophone of \data{u} before \data{k}.
Only the case of glottal stop needs to be considered.  So we revise
the form, replacing \verb|$V| with just the vowels in question, and
replacing the \verb|$C| of the coda with apostrophe (for glottal
stop).  We add a term for the syllable onset and resubmit the query.
See Figure~\ref{srch:hrv}.
This time, several attributes are omitted from the display for brevity.

\begin{dpy}
  \begin{query}
    display: & count\\
    root:    & .*([\$C]+)([ou])'\# \\
    axes:    & flip
  \end{query}
  \begin{results}{c|ccccccccccccccccccc}
    &w&p&pf&b&t&ts&d&c&j&k&g&f&v&s&z&m&n&\eng&l\\ \hline
    u &6&8&1 &1&6&  &1&6&4&5&3&5&2& & &4&1&1   &5\\
    o & & &1 & & &6 & & & & & & &1&1&3& & &    &
  \end{results}
  \caption{Query to Probe the Phonemic Status of the O/U Contrast}
  \label{srch:hrv}
\end{dpy}

We can now conclude that \data{o} and \data{u} are in complementary distribution,
except for the five words corresponding to \data{pf} and \data{v} onsets.
But what are these words?
We revise the form again, further restricting the search string as
follows:\spc

\begin{query}
  display: & speech word gloss\\
  root:    & .*(pf$\mid$v)[ou]'\#
\end{query}\spc

The display parameter is set to
{\small \verb|speech word gloss|}
allowing us to see (and hear) the
individual lexical items.
The results are shown below.\spc

\begin{results}{l|l}
  pf & \speech \data{lepfo'} {\it mortar}\\
  & \speech \data{mpfu'} {\it blood pact}\\ \hline
  v  & \speech \data{mvo'} {\it space in front of bed}\\
  & \speech \data{avu'} {\it remainder}\\
  & \speech \data{levu't\'e} {\it kitchen woodpile}\\
\end{results}\spc

\noindent
The cells of the output table now contain fragments of the
lexical entries.  The first part is an icon which, when clicked,
plays the speech file.  The second part is a gif of the orthographic
form of the word.  The third part is the English gloss.
Note that the above nouns have different prefixes
(e.g.~\data{le-}, \data{m-}, \data{a-}).  These are noun class prefixes and
are not part of the \verb|root| field.  If we had wanted to take
prefixes into consideration then the \verb|as| attribute, containing a
transcription of the whole word, could have been used instead.

Listening to the speech files it was found that the syllables
\data{pfo'} and \data{pfu'} sounded exactly the same, as did
\data{vo'} and \data{vu'}.
The whole process up to this point had taken less than five minutes.
After some quick informant work to recheck the data and
hear the native-speaker intuitions, it was clear that the
distinction bet\-ween \data{o} and \data{u} in closed syllables was
subphonemic.

\section{MORE POWERFUL QUERIES}
\label{sec:queries}

\subsection{Constraining one field and displaying another}
\label{sec:split}

In some situations we are not interested in seeing the field which
was constrained, but another one instead.  The next query displays the tone
field for monosyllabic roots, classed into
open and closed syllables.  Although the \verb|root| attribute is used
in the query, the \verb|root| field is not actually displayed.
(This query makes use of a projection function
which maps all consonants onto \verb|C| and all vowels onto \verb|V|,
as will be explained later.)\spc

\begin{query}
  display:    & tone \\
  root:       & \#C+V(C?)\# (\$CV-proj)
\end{query}\spc

\noindent
The \verb|C+| expression denotes a sequence of one or more consonants,
while \verb|C?| denotes an optional coda
consonant.  By making \verb|C?| into a parameter (using parentheses)
the search results will be presented in a two column table,
one column for open syllables (with a null label)
and one for closed syllables (labelled \verb|C|).
A minor change to the \verb|root| attribute, enlarging the
scope of the parameter ({\small \verb|\#C+(VC?)\#|}), will
produce the more satisfactory column labels \verb|V| and \verb|VC|.

\subsection{Searching for near-minimal sets}

Finding good minimal sets is a heuristic process.  No attempt has
been made to encode heuristics into the system.  Rather, the aim has
been to permit flexible interaction between user and system as a
collection of minimal sets is refined.
To facilitate this process, the regular expression notation is
extended slightly.  Recall the way that parameters (parenthesised
subexpressions) allowed output to be structured.  One of the
parameters will be said to be {\em in focus}.  Syntactically,
this is expressed using braces instead of parentheses.  Semantically,
such a parameter becomes the focus of a search for minimal sets.
Typically, this parameter will contain a list of segments, such
as \verb|{[ou]}|, or an optional segment whose presence is to be
contrasted with its absence, such as \verb|{h?}|.

In order for a minimal set to be found,
the parameter in focus must have more than one possible instantiation,
while the other parameters remain unchanged.  To see how this
works, consider the following example.  Suppose we wish to identify the
minimal pairs for \data{o/u} discussed above, but without having to specify
glottal stop in the query, as shown in Figure~\ref{srch:minou}.
Note this example of a 3D table.

\begin{dpycol}
  \begin{query}
    display: & word gloss\\
    root:    & .*([\$C]+)\{[ou]\}([\$C])\#
  \end{query}
  \begin{results}{l|l}
    & ' \\ \hline \\
    pf & \fbox{\begin{tabular}{l}lepfo' {\it mortar}\\ \hline
        mpfu' {\it blood pact} \end{tabular}} \\ \\
    v  & \fbox{\begin{tabular}{l}mvo' {\it space in front of bed}\\ \hline
        avu' {\it remainder} \\
        levu't\'{\i} {\it kitchen woodpile}\end{tabular}}
  \end{results}
  \caption{Minimal Sets for O/U}
  \label{srch:minou}
\end{dpycol}

If this was not enough minimal pairs, we could relax the restrictions
on the context.  For example, if we do not wish to insist on the
following consonant being identical across minimal pairs, we can
remove the second set of parentheses thus:
{\small \verb|.*([$C]+){[ou]}[$C]#|}.  This
now gives minimal pairs like
\data{leg\=ok} {\it work} and \data{\eng gu'} {\it year}.  Observe that the
consonant preceding the \data{o/u} vowel is fixed across the minimal
pair, since this was still parenthesised in the query string.

Usually, it is best for minimal pairs to have similar syntactic
distribution.  We can add a restriction that all minimal pairs
must be drawn from the same syntactic category by making the whole
\verb|part| attribute into a parameter as follows.\spc

\begin{query}
  display:    & tone \\
  root:    & .*([\$C]+)\{[ou]\}[\$C]\# \\
  part:       & (.*)
\end{query}\spc

\noindent
Making the \verb|part| attribute into a parameter adds an extra
dimension to the table of results.  We now only see an \data{o/u}
minimal pair if the other parameters agree.  In other words,
all minimal pairs that are reported will contain the same consonant
cluster before the \data{o/u} vowel and will be from the same
syntactic category.

\subsection{Variables across attributes}

There are occasions where we need to have the same variable appearing
in different attributes.  For example, suppose we wanted to check where
the southern dialect and the principal dialect have identical vowels:\footnote{
  Roots are virtually all monosyllabic, so there will usually be a
  unique vowel sequence for the {\tt [\$V]+} in the
  regular expression to match with.
}\spc

\begin{query}
  display:    & root s\_dialect \\
  root:       & .*(3[\$V]+).* \\
  s\_dialect: & .*\$3.*
\end{query}\spc

\noindent
This query makes use of another syntactic extension to regular
expressions.  An arbitrary one-digit number which appears immediately
inside a parameter allows the parameter to be referred to elsewhere.
This means that whichever sequence of vowels matches \verb|[$V]+| in
the \verb|root| field must also appear somewhere in the
\verb|s_dialect| field.

\subsection{Negative restrictions}

The simplest kind of negative restriction is built using the set
complement operator (the caret).
However this only works for single character complements.  A much
more powerful negation is available with the \verb|?!| zero-width
negative lookahead assertion, available in Perl 5, which I will
now discuss.

The next example uses the tone attribute.  Dschang is a tone language,
and the records in the lexicon include a field containing a tone melody.
Tone melodies consist of the characters H (high), L (low), D
(downstep) and F (fall).  A single tone has the form
\verb|D?[HL]F?|, i.e. an optional downstep, followed by
\verb|H| or \verb|L|, followed by an optional fall.
The next example finds all entries starting with a sequence of
unlike tones.\spc

\begin{query}
  root:       & .*(1[\$T])(?!\$1)[\$T].* \\
  vars:       & \$T = D?[HL]F?
\end{query}\spc

\noindent
The \verb|(1[$T])| expression matches any tone and sets the
\verb|$1| variable to the tone which was matched.  The
\verb|(?!$1)| expression requires that whatever follows
the first tone is different, and the final \verb|[$T]| insists
that this same following material is a tone (rather than being
empty, for example).\footnote{Care must be taken to ensure
that the alphabetic encodings of distinct tones are sufficiently
different from each other, so that one is not an initial substring
of another.}

\subsection{Projections}
\label{sec:projections}

I earlier introduced the notion of projections.  In fact,
the system allows the user to
apply an arbitrary manipulation to any attribute before the
matching is carried out.  Here is the query again, this time
with the {\small \verb|$CV-proj|} variable filled out.\spc

\begin{query}
  display:    & tone \\
  root:       & \#C+V(C?)\# (\$CV-proj) \\
  vars:       & \$CV-proj = \{tr/\$C/C/; tr/\$V/V/;\}
\end{query}\spc

\noindent
This causes the Perl \verb|tr| (transliterate) function to be applied to
the \verb|root| attribute before the \verb|#C+V(C?)#| regular expression
is matched on this field.

Projections can also be used to simulate second order variables,
such as required for place of articulation.  Suppose that the language
has three places of articulation: L (labial), A (alveolar) and V (velar).
We are interested in finding any unassimilated sequences in the data
(i.e.~adjacent consonants with different places of articulation).
The following query does just this.  Prior to matching, the segments
which have a place of articulation value are projected to that value,
again using {\small \verb|tr|}.
The query expression looks for a sequence of
any pair \verb|$P$P|, where \verb|$P| is a second order
variable ranging over places of articulation.\spc

\begin{query}
  display:    & word \\
  root:       & .*(5\$P)(?!\$5)(\$P).* (\$P-proj) \\
  vars:       & \$P-proj=tr/pbmtdnkgN/LLLAAAVVV/; \\
              & \$P = [LAV];
\end{query}\spc

Observe that the second \verb|$P| must be different from the first,
because of the zero-width negative lookahead assertion
\verb|(?!$5)|.  This states that immediately to the right of this
position one does not find an instance of \verb|$5|, where this
variable is the place of articulation found in the first position.
The output of the query is a $3 \times 3$ table showing all words that contain
unassimilated consonant sequences.

\section{SYSTEM OVERVIEW}
\label{sec:system}

\subsection{Lexicon compiler}

The base lexicon is in Shoebox format, in which the fields are not
required to be in a fixed order.  To save on runtime processing, a
preprocessing step is applied to each field.  For example, the contents of the
\verb|\w| field, comprising characters from the Cameroon character
set, are replaced by a pointer a graphics file for the word
(i.e.~a URL referencing a gif).\footnote{These gifs were
  generated using \LaTeX\ along with
  the utilities {\small \verb|pstogif|} and {\small \verb|giftool|}.}
Each record is
processed into a single line, where fields occur in a canonical order
and a field separator is inserted, and the compiled lexicon is stored
as a DBM file for rapid loading.

\subsection{The query string}
\label{sec:query}

The search attributes in the query form can contain arbitrary Perl V5 regular
expressions, along with some extensions introduced above.
A CGI program constructs a query string based on the submitted form
data.  The query string is padded with wild cards for those fields
which were not restricted in the query form.

The dimensionality of the output and the axis labels are determined by
the appearance of `parameters' in the search attributes.  These
parenthesised subexpressions are copied
directly into the query string.  So, for
example, the first query above contained the search expression
{\small \verb|.*([$V])([$C])#|} applied to the \verb|root| field.  This
field occupies fifth position in the compiled version of a record, and
so the search string is as follows.  The variable \verb|$e| matches
any sequence of characters not containing the field separator.

\noindent
{\small
\begin{tabular}{ll}
\verb|$search =| & \verb|/^$e;$e;$e;$e;.*([$V])([$C])#;|\\
                 & \verb|$e;$e;$e;$e;$e;$e;$e;$e$/|
\end{tabular}}\spc

\subsection{The search loop}

Search involves a linear pass over the whole
lexicon \verb|%LEX|.\footnote{Inverting on
  individual fields was avoided because
  of the runtime overheads and the fact that this prevents variable
  instantiation across fields.}
The parameters contained in \verb|$search| are tied to the variables
\verb|$1| -- \verb|$4|.  These are stored in four associative arrays
\verb|$dim1| -- \verb|$dim4| to be used later as axis labels.


{\small
\begin{verbatim}
foreach $entry (keys %LEX) {
  if ($LEX{$entry} =~ /$search/) {
    $dim1{$1}++;
    $dim2{$2}++;
    $dim3{$3}++;
    $dim4{$4}++;
    $hits{"$1;$2;$3;$4"} .= ";".$entry;
  }
}
\end{verbatim}}\spc

Finally, a pointer to the entry is stored in the 4D array
\verb|$hits| (appended to any existing hits in that cell.)
Here we see that the structuring of the output table
using parameters is virtually transparent, with Perl itself
doing the necessary housekeeping.

As an example, suppose that the following lexical entry is being
considered at the top of the above loop:

\noindent
{\small
\begin{tabular}{l}
\verb|$entry =| \verb|0107|\\
\verb|$LEX{$entry} =|\\
\hspace*{1em}\verb|0107; ;<img src="akup.gif">;|\\
\hspace*{1em}\verb|#a.kup#;#kup#;LL;;*k`ub`;n;7/6,8;|\\
\hspace*{1em}\verb|skin, bark;peau,\'ecorce;|
\end{tabular}}\spc

\noindent
By matching this against the query string given in our first example we
end up matching \verb|.*([$V])([$C])#| with \verb|#kup#|.  This
results in \verb|$1=u| and \verb|$2=p|.  The entries
\verb|$dim1{u}| and \verb|$dim2{p}| are incremented, recording
these values for later use in the \verb|$V| and \verb|$C| axes respectively.
Finally \verb|$hits("u;p;;")| is updated with the index \verb|0107|.

\subsection{The display loop}

This module cycles through the axis labels that were stored in
\verb|$dim1| -- \verb|$dim4| and combines them to access the
\verb|$hits| array.  At each level of nesting, code is generated
for the \HTML\ or \LaTeX\ table output.  At the innermost level,
the fields selected by the user in the
\verb|display| attribute are used to build the current cell.

\section{FUTURE WORK}

A number of extensions to the system are planned.  Since Dschang is
a tone language, it would be particularly valuable to have access to
the pitch contours of each word.  These will eventually be displayed
as small gifs, attached to the lexical entries.

Another extension would be to permit updates to the lexicon
through a forms interface.  A special instance of the search form
could be used to validate existing and new entries, alerting the
user to any data which contradicts current hypotheses.

The regular expression notation is sometimes cumbersome and opaque.
It would be useful to have a higher level language as well.
One possibility is the notation of autosegmental phonology,
which can be compiled into finite-state automata \cite{BirdEllison94}.
The graphics capabilities for this could be provided on the client
side by a Java program.

A final extension, dependent on developments with \HTML\ itself,
would be to provide better support for special
characters and user-definable keystrokes for accessing them.

\section{CONCLUSION}

This paper has presented a hypertext lexicon tailored to the practical
needs of the phonologist working on large scale data problems.  The
user accesses the lexicon via a forms interface provided by \HTML\ and
a browser.  A CGI program processes the query.  The user can refine
a query during the course of several interactions with the system,
finally switching the output to \LaTeX\ format for direct inclusion
of the results in a research paper.  An extension to
the regular expression notation was used for searching for minimal pairs.
Parenthesised subexpressions are interpreted as parameters which
control the structuring of search results.  These extensions, though
intuitively simple, make a lot of expressive power available to the user.
The current prototype system has been used heavily for 
substantive phonological fieldwork and analysis on the field,
documented in \cite{Bird97syl}.
There are a number of ensuing
benefits of this approach for phonological research:
(i) it supports a quantitative method of doing phonological research;
(ii) it gives universal access to the same set of informants;
(iii) it enables other researchers to hear the
original speech data without having to rely on published transcriptions;
(iv) it makes the full power of regular expression search
available, and search results are full multimedia documents; and
(v) it enables the early refutation of false hypotheses, shortening
the analysis-hypothesis-test loop.

\section*{ACKNOWLEDGEMENTS}

This research is funded by the
the UK Economic and Social Research Council, under grant R00023 5540
{\it A Computational Model of Tone and its Relationship to Speech}.
My activities in Cameroon were covered by a research permit with the Ministry of
Scientific and Technical Research of the Cameroon government,
number 047/MINREST/D00/D20.  I am grateful to Dafydd Gibbon for
helpful comments on an earlier version of this paper.


\begin{thebibliography}{}

\bibitem[\protect\citeauthoryear{Bevan}{Bevan}{1995}]{Findphone}
Bevan, D. (1995).
\newblock {\em FindPhone User's Guide: Phonological Analysis for the Field
  Linguist, Version 6.0}.
\newblock Waxhaw NC: SIL.

\bibitem[\protect\citeauthoryear{Bird}{Bird}{1997}]{Bird97syl}
Bird, S. (1997).
\newblock Dschang Syllable Structure.
\newblock In H.~{van der Hulst} \& N.~Ritter (Eds.), {\em The Syllable: Views
  and Facts}. Oxford University Press.
\newblock To appear.

\bibitem[\protect\citeauthoryear{Bird \& Ellison}{Bird \&
  Ellison}{1994}]{BirdEllison94}
Bird, S. \& Ellison, T.~M. (1994).
\newblock One level phonology: autosegmental representations and rules as
  finite automata.
\newblock {\em Computational Linguistics}, {\em 20}, 55--90.

\bibitem[\protect\citeauthoryear{Bird \& Tadadjeu}{Bird \&
  Tadadjeu}{1997}]{BirdTadadjeu97}
Bird, S. \& Tadadjeu, M. (1997).
\newblock {\em Petit Dictionnaire Y\'emba-Fran\c{c}ais (Dschang-French
  Dictionary)}.
\newblock Cameroon: ANACLAC.

\bibitem[\protect\citeauthoryear{Buseman, Buseman \& Early}{Buseman
  et~al.}{1996}]{Shoebox}
Buseman, A., Buseman, K., \& Early, R. (1996).
\newblock {\em The Linguist's Shoebox: Integrated Data Management and Analysis
  for the Field Linguist}.
\newblock Waxhaw NC: SIL.

\bibitem[\protect\citeauthoryear{Coleman, Dirksen, Hussain \& Waals}{Coleman
  et~al.}{1996}]{ColemanDirksen96}
Coleman, J., Dirksen, A., Hussain, S., \& Waals, J. (1996).
\newblock Multilingual phonological analysis and speech synthesis.
\newblock In {\em Computational Phonology in Speech Technology: Proceedings of
  the Second Meeting of the ACL Special Interest Group in Computational
  Phonology}, (pp.\ 67--72). Association for Computational Linguistics.

\bibitem[\protect\citeauthoryear{Ellison}{Ellison}{1992}]{Ellison92b}
Ellison, T.~M. (1992).
\newblock {\em Machine Learning of Phonological Structure}.
\newblock PhD thesis, University of Western Australia.

\bibitem[\protect\citeauthoryear{Kaplan \& Kay}{Kaplan \&
  Kay}{1994}]{KaplanKay94}
Kaplan, R.~M. \& Kay, M. (1994).
\newblock Regular models of phonological rule systems.
\newblock {\em Computational Linguistics}, {\em 20}, 331--78.

\bibitem[\protect\citeauthoryear{Kornai}{Kornai}{1995}]{Kornai95}
Kornai, A. (1995).
\newblock {\em Formal Phonology}.
\newblock New York: Garland Publishing.

\bibitem[\protect\citeauthoryear{Lowe \& Mazaudon}{Lowe \&
  Mazaudon}{1994}]{LoweMazaudon94}
Lowe, J.~B. \& Mazaudon, M. (1994).
\newblock The Reconstruction Engine: a computer implementation of the
  comparative method.
\newblock {\em Computational Linguistics}, {\em 20}, 381--417.

\bibitem[\protect\citeauthoryear{Reinhard \& Gibbon}{Reinhard \&
  Gibbon}{1991}]{Reinhard91}
Reinhard, S. \& Gibbon, D. (1991).
\newblock Prosodic inheritance and morphological generalizations.
\newblock In {\em Proceedings of the Fifth Conference of the European Chapter
  of the Association for Computational Linguistics}, (pp.\ 131--6). Association
  for Computational Linguistics.

\bibitem[\protect\citeauthoryear{Wall \& Schwartz}{Wall \&
  Schwartz}{1991}]{Wall91}
Wall, L. \& Schwartz, R.~L. (1991).
\newblock {\em Programming Perl}.
\newblock O'Reilly and Associates.

\end{thebibliography}
\end{document}